\documentclass[sigconf, screen]{acmart}
\AtBeginDocument{%
  }

\usepackage{algorithm}
\usepackage{algpseudocode}
\begin{document}

\title{Fast-then-Fine: A Two-Stage Framework with Multi-Granular Representation for Cross-Modal Retrieval in Remote Sensing}

\author{Xi Chen}
\email{2024102110094@whu.edu.cn}
\affiliation{%
	\institution{School of Computer Science, Wuhan University}
	\city{Wuhan}
	\postcode{430072}
	\country{China}
}

\author{Xu Chen}
\email{xuchen@whu.edu.cn}
\affiliation{%
	\institution{School of Computer Science, Wuhan University}
	\city{Wuhan}
	\postcode{430072}
	\country{China}
}

\author{Xiangyang Jia}
\email{jxy@whu.edu.cn}
\affiliation{%
	\institution{School of Computer Science, Wuhan University}
	\city{Wuhan}
	\postcode{430072}
	\country{China}
}

\author{Xu Zhang}
\email{zhangx0802@whu.edu.cn}
\affiliation{%
	\institution{School of Computer Science, Wuhan University}
	\city{Wuhan}
	\postcode{430072}
	\country{China}
}

\author{Shuquan Wei}
\email{2021302111045@whu.edu.cn}
\affiliation{%
	\institution{School of Computer Science, Wuhan University}
	\city{Wuhan}
	\postcode{430072}
	\country{China}
}

\author{Wei Wang}
\email{wangwei@bigai.ai}
\affiliation{%
	\institution{Beijing Institute for General Artificial Intelligence (BIGAI)}
	\city{Beijing}
	\postcode{100080}
	\country{China}
}

\renewcommand{\shortauthors}{X. Chen et al.}

\begin{abstract}
  Remote sensing (RS) image–text retrieval plays a critical role in understanding massive RS imagery. However, the dense multi-object distribution and complex backgrounds in RS imagery make it difficult to simultaneously achieve fine-grained cross-modal alignment and efficient retrieval.
  Existing methods either rely on complex cross-modal interactions that lead to low retrieval efficiency, or depend on large-scale vision–language model pre-training, which requires massive data and computational resources. To address these issues, we propose a fast-then-fine (FTF) two-stage retrieval framework that decomposes retrieval into a text-agnostic recall stage for efficient candidate selection and a text-guided rerank stage for fine-grained alignment. Specifically, in the recall stage, text-agnostic coarse-grained representations are employed for efficient candidate selection; in the rerank stage, a parameter-free balanced text-guided interaction block enhances fine-grained alignment without introducing additional learnable parameters. Furthermore, an inter- and intra-modal loss is designed to jointly optimize cross-modal alignment across multi-granular representations. Extensive experiments on public benchmarks demonstrate that the FTF achieves competitive retrieval accuracy while significantly improving retrieval efficiency compared with existing methods.
  
\end{abstract}




\maketitle

\section{Introduction}
Remote sensing (RS) imagery has grown rapidly with the continuous development of earth observation platforms, generating massive volumes of visual data~\cite{Wang_2025_ICCV,feng2026earthagent,Astruc_2025_CVPR}.  Efficient access to and understanding of RS imagery have become increasingly important~\cite{li2025lhrs,soni2025earthdial}. 
Image–text retrieval provides an effective way to retrieve RS imagery through textual queries by linking visual content with semantic descriptions~\cite{23acmmmPIR,23tgrsKAMCL,25tgrsCLGSA,24acmmmEBAKER}. 
However, RS imagery typically exhibits high spatial resolution and complex background composition, where diverse objects and fine structural details coexist within a single scene, making it difficult to simultaneously achieve efficient retrieval and fine-grained cross-modal alignment.

\begin{figure}[!htb]
	\centering
	\includegraphics[width=\columnwidth]{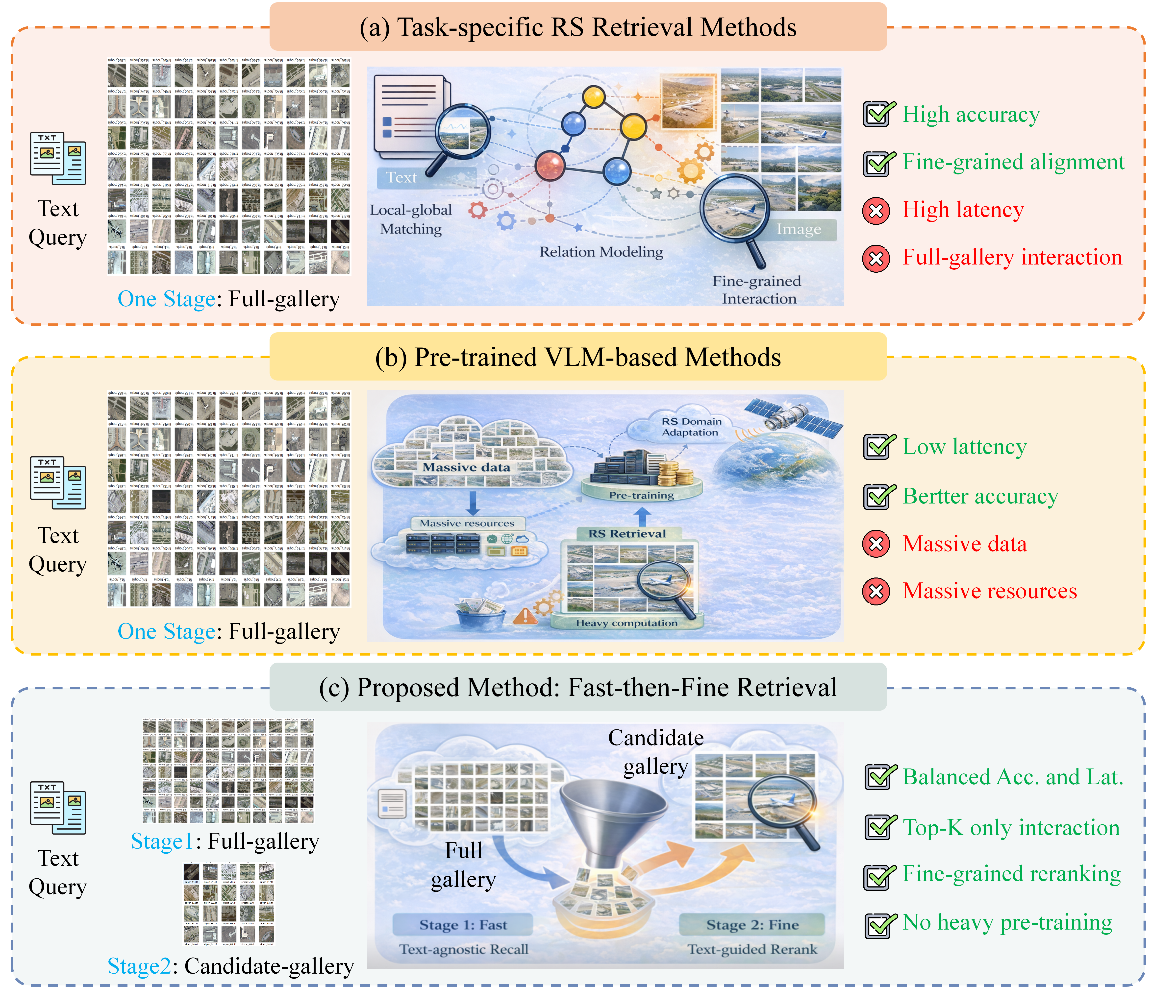}
	\caption{
	Comparison of three paradigms for remote sensing image--text retrieval.
	(a) Task-specific methods perform fine-grained matching via complex full-gallery interaction, but suffer from high latency.
	(b) Pre-trained VLM-based methods achieve strong generalization, but require large-scale data and heavy computation.
	(c) Our fast-then-fine framework performs full-gallery recall followed by candidate-gallery reranking, achieving a better balance between retrieval accuracy and efficiency without large-scale pre-training.
	}
	\label{fig:motivation}
\end{figure}

\begin{figure}[!htb]
	\centering
	\includegraphics[width=\columnwidth]{figures/intro.png}
	\caption{Accuracy--efficiency comparison of task-specific RS image--text retrieval methods. FTF achieves more favorable operating points under different candidate sizes $K$, showing a better balance between retrieval accuracy and efficiency without relying on large-scale VLM pre-training.}
	\label{fig:intro_tradeoff}
\end{figure}

Recent years have witnessed substantial progress in RS image–text retrieval~\cite{25acmmmEAPA}. On one hand, many methods leverage complex cross-modal interactions to achieve fine-grained alignment between RS imagery and text~\ref{fig:motivation}(a), which improves retrieval accuracy but often introduces significant computational overhead and limits retrieval efficiency~\cite{22tgrsGaLR,23sensorsFAAMI,24tgrsMSCMAT,24tgrsIERR}. On the other hand, recent advances in vision–language model (VLM) pre-training further improve retrieval performance by leveraging large-scale image–text data~\ref{fig:motivation}(b), but these approaches typically require massive training data and considerable computational resources~\cite{23tgrsPETL,24aaaiskysrcipt,24iclrGRAFT,24tgrsGeoRSCLIP}. 
As a result, achieving a practical balance between retrieval accuracy and efficiency remains a critical challenge for RS image–text retrieval. As illustrated in Fig.~\ref{fig:intro_tradeoff}, existing task-specific methods often exhibit a clear balance between retrieval accuracy and retrieval efficiency, while the proposed FTF achieves more favorable balancing points~\cite{22tgrsGaLR,23acmmmPIR,25tgrsFSSN}.

To address the above challenges, we propose a fast-then-fine (FTF) two-stage retrieval framework for RS image–text retrieval~\ref{fig:motivation}(c). The key idea is to decouple retrieval into an efficient candidate selection stage and a fine-grained alignment stage. Specifically, the framework first performs text-agnostic recall using coarse-grained representations to rapidly identify a small set of candidate RS imagery. It then performs text-guided rerank to refine the semantic correspondence between RS imagery and textual queries. This design enables the framework to reduce expensive cross-modal interactions over the entire database while preserving the ability to achieve fine-grained cross-modal alignment. As a result, the proposed framework provides an effective solution for balancing retrieval accuracy and efficiency in RS image–text retrieval.

The main contributions of this work are summarized as follows:
\begin{itemize}
	\item We propose a FTF two-stage retrieval framework for RS image--text retrieval, which decomposes retrieval into a text-agnostic recall stage and a text-guided rerank stage to balancing retrieval efficiency and accuracy.
	
	\item We introduce a multi-granular representation learning strategy with a parameter-free balanced text-guided interaction block (BTIB), which enhances the semantic correspondence between RS imagery and textual queries without introducing additional learnable parameters.
	
	\item We design an inter- and intra-modal (I$^{2}$M) loss to jointly optimize cross-modal alignment across multi-granular representations, improving representation consistency.
	
	\item Extensive experiments on public benchmarks demonstrate that the proposed method achieves competitive retrieval accuracy while significantly improving retrieval efficiency compared with existing methods.
\end{itemize}

\section{Related Work}
\subsection{Task-specific RS Retrieval Methods.} Existing RS image--text retrieval methods mainly improve retrieval accuracy by designing increasingly complex cross-modal interaction mechanisms, thereby enhancing semantic alignment between RS imagery and textual descriptions. Representative methods, including CAMP~\cite{2019iccvCAMP}, LW-MCR~\cite{21tgrsLW-MCR}, AMFMN~\cite{22tgrsAMFMN}, KAMCL~\cite{23tgrsKAMCL}, and SCAN~\cite{2018eccvSCAN}, focus on establishing semantic correspondence through cross-modal projection, attention-based matching~\cite{25patrecCMA,25cvprCAYN}, and semantic correlation modeling~\cite{25nnUSSL,25airCSAN}, which lay an important foundation for RS image--text retrieval. 
Building on this line, more recent methods, such as GaLR~\cite{22tgrsGaLR}, PIR~\cite{23acmmmPIR}, FAAMI~\cite{23sensorsFAAMI}, MTGFE~\cite{23RemoteSensingMTGFE}, MSCMAT~\cite{24tgrsMSCMAT}, IERR~\cite{24tgrsIERR}, VGSGN~\cite{24tgrsVGSGN}, SIPN~\cite{25tgrsSIPN}, and FSSN~\cite{25tgrsFSSN}, further strengthen fine-grained cross-modal alignment through local-to-global matching~\cite{25tgrsCLGSA,25cvprGOAL}, relation-aware modeling~\cite{25cvprRAA,25tpamiSRA}, multi-granular feature learning, and richer semantic interaction~\cite{23sigirMGRL,25neunetSFE,24asocCCCL}. 
These methods indicate that fine-grained modeling is crucial for handling the high spatial resolution and complex background characteristics of RS imagery~\cite{Asokan_2025_CVPR,ma2025multimodal}. However, these methods apply complex cross-modal interaction directly over the full candidate space, which limits retrieval efficiency in practical retrieval scenarios.

\subsection{Pre-trained VLM-based methods.} In addition to task-specific RS image--text retrieval methods, recent studies have further improved retrieval performance by leveraging large-scale VLM pre-training~\cite{21icmlCLIPBASE,22icmlBLIP,23icmlBLIP2}. 
Representative methods, including CLIP-Base~\cite{21icmlCLIPBASE} and PETL~\cite{23tgrsPETL}, transfer general-purpose pre-trained vision--language representations to the RS domain, while SkyScript~\cite{24aaaiskysrcipt}, GRAFT~\cite{24iclrGRAFT}, RemoteCLIP~\cite{24tgrsremoteclip}, and GeoRSCLIP~\cite{24tgrsGeoRSCLIP} further enhance retrieval accuracy through RS-oriented pre-training or domain adaptation. 
These methods demonstrate that large-scale pre-training can significantly strengthen cross-modal representation learning and improve semantic alignment~\cite{25cvprPostAlign,24eccvSILC,25tpamiGLSCL,24prModalInteraction} between RS imagery and textual descriptions. 
However, their performance usually depends on large-scale image--text pairs and substantial computational resources for pre-training or adaptation~\cite{yamaguchi2025postpretraining,wei2025hqclip,huang2025noiserobust}. 
As a result, although VLM pre-training improves the retrieval upper bound, it does not directly address how to achieve efficient and fine-grained retrieval under limited training data and computational resources~\cite{24eccvSILC,25cvprPostAlign,23tgrsPETL}.

\subsection{Efficient cross-modal retrieval.} 
Based on the above observations, existing methods improve matching accuracy by relying on increasingly sophisticated cross-modal interaction, and their computational cost grows substantially when such interaction is applied to the full candidate space. Therefore, a practical retrieval framework should jointly consider efficient candidate selection and fine-grained semantic refinement~\cite{24acmmmCREAM,25aclOMGM,25prCrossTVR}. 
This motivates a two-stage retrieval design~\cite{24sigirCFIR,25aclOMGM,25prCrossTVR,25ipmAFRRank}, in which a lightweight recall stage first narrows the search space, followed by a rerank stage that further refines semantic correspondence over a small set of candidates. Such a design provides a natural way to balancing retrieval efficiency and accuracy, yet remains insufficiently explored in RS image--text retrieval~\cite{glar2022,cmpagl2024,tsr2017,dai2025rerank}.

\section{Methods}

\subsection{Problem Formulation}

Let $\mathcal{D} = \{(I_i, x_i)\}_{i=1}^{N}$ denote a training set of paired RS images $I_i$ and textual descriptions $x_i$, where $N$ is the number of samples.  
Given a textual query $x_q$, the goal of RS image--text retrieval is to return a ranked list of images from a large gallery $\mathcal{G}$ that are semantically relevant to $x_q$.  

Each textual description $x_i$ is encoded into a token-level representation $T_i$ and a global text embedding $t_i$ using a text encoder.  
Each image $I_i$ is represented by three types of visual embeddings:  
\begin{itemize}
	\item $\mathbf{v}_1$: text-agnostic embedding for recall, obtained by multi-granular visual encoding;  
	\item $\mathbf{v}_2$: text-guided coarse embedding for rerank;  
	\item $\mathbf{v}_3$: text-guided fine embedding for rerank.  
\end{itemize}

The overall objective is to learn embeddings $\mathbf{v}_1, \mathbf{v}_2, \mathbf{v}_3$ and $t$ such that (i) $\mathbf{v}_1$ can efficiently identify a small set of candidate images in the recall stage, and (ii) $\mathbf{v}_2, \mathbf{v}_3$ enable fine-grained semantic alignment with the query in the rerank stage.  
For both text and visual encoding, we adopt transformer-based encoders. Detailed backbone settings are provided in the experimental section.

\subsection{Overview}

We denote the gallery of RS images as $\mathcal{G}$ and a textual query as $x_q$.  
The FTF framework aims to retrieve a ranked subset of candidate images $\mathcal{C} \subset \mathcal{G}$ that are semantically relevant to $x_q$ while maintaining high efficiency.  
Formally, the recall stage computes a text-agnostic mapping
\begin{equation}
	\mathbf{v}_1 = f_{\mathrm{recall}}(I), \quad I \in \mathcal{G},
	\label{eq:recall_mapping}
\end{equation}
to produce multi-granular embeddings for efficient candidate selection, where $f_\mathrm{recall}$ aggregates coarse and fine patch token embeddings.  
The rerank stage then performs a text-guided refinement
\begin{equation}
	\mathbf{v}_2, \mathbf{v}_3 = f_{\mathrm{rerank}}(\mathbf{v}_1, x_q),
	\label{eq:rerank_mapping}
\end{equation}
where $\mathbf{v}_2$ and $\mathbf{v}_3$ correspond to coarse and fine text-guided embeddings generated by BTIB.  

\begin{figure*}[!htb]
	\centering
	\includegraphics[width=1.0\linewidth]{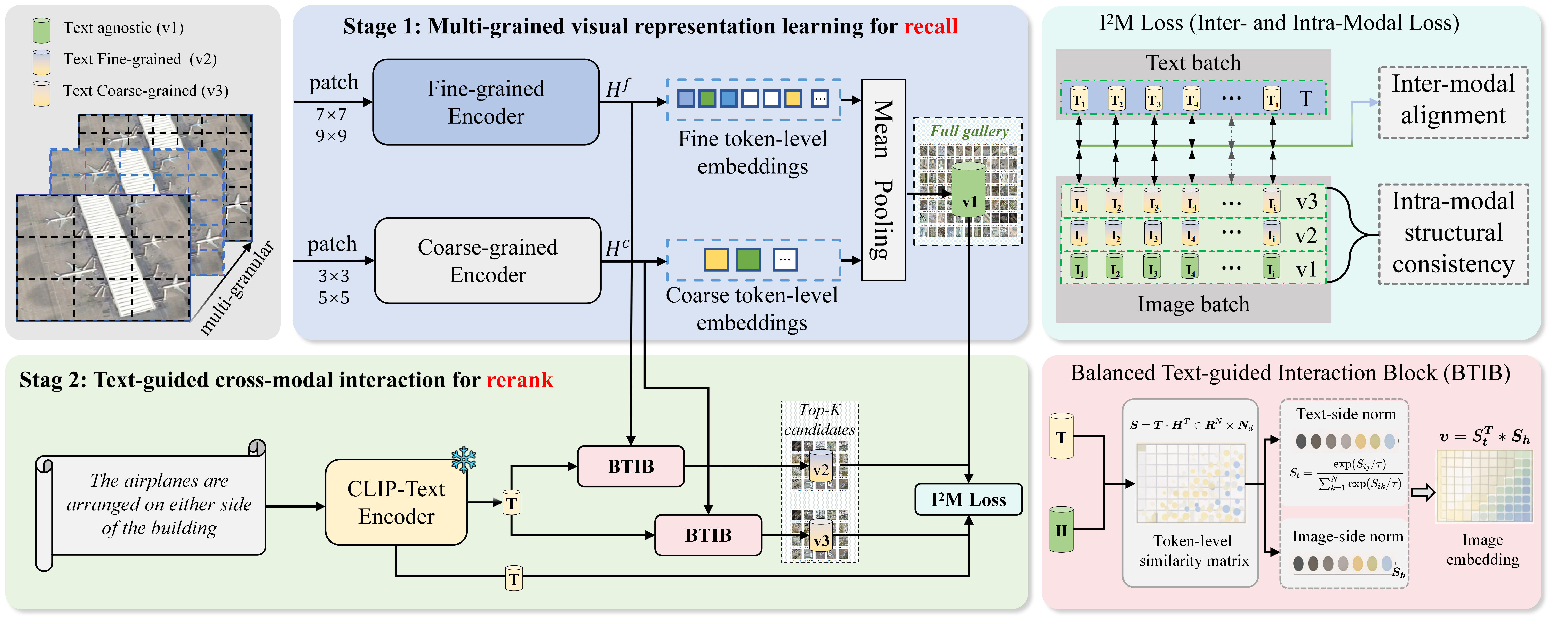}
	\caption{The overall FTF two-stage retrieval framework. 
		Multi-granular visual embeddings $\mathbf{v}_1$ are used for text-agnostic recall, 
		while BTIB generates $\mathbf{v}_2$ and $\mathbf{v}_3$ in the text-guided rerank stage. 
		I$^2$M Loss jointly optimizes cross-modal alignment and intra-modal structural consistency.}
	\label{fig:framework}
\end{figure*}

Figure~\ref{fig:framework} illustrates the overall framework.  
Given a query, the recall stage first selects a small set of candidates using $\mathbf{v}_1$, avoiding expensive text-conditioned interaction over the full gallery.  
The rerank stage then refines the candidate ranking with $\mathbf{v}_2$ and $\mathbf{v}_3$, where BTIB provides parameter-free text-guided interaction to achieve fine-grained semantic alignment.  
This two-stage design provides a natural way to balancing retrieval efficiency and retrieval accuracy, as the expensive text-guided interaction is only applied to a small candidate set and uses the parameter-free BTIB module.

\begin{algorithm}[t]
\caption{Fast-then-Fine Retrieval with BTIB}
\label{alg:ftf}
\begin{algorithmic}[1]
	\Require Gallery $\mathcal{G}$, query text $x_q$, candidate size $K$
	\Ensure Ranked candidate list
	
	\Statex \textbf{Stage 1: Text-agnostic Recall}
	\For{each image $I \in \mathcal{G}$}
	\State Extract multi-granular tokens $\mathbf{H}^c, \mathbf{H}^f$
	\State Compute recall embedding $\mathbf{v}_1 = \mathrm{Agg}(\mathbf{H}^c, \mathbf{H}^f)$
	\EndFor
	\State $\mathcal{C} \leftarrow \mathrm{TopK}(\cos(t_q, \mathbf{v}_1), K)$
	
	\Statex \textbf{Stage 2: Text-guided Rerank}
	\For{each candidate $I \in \mathcal{C}$}
	\State Compute $\mathbf{v}_2, \mathbf{v}_3 = \mathrm{BTIB}(\mathbf{H}^c, \mathbf{H}^f, T_q)$
	\State $s(I) \leftarrow (1-\lambda)\cos(t_q, \mathbf{v}_2) + \lambda \cos(t_q, \mathbf{v}_3)$
	\EndFor
	
	\State \Return candidates in $\mathcal{C}$ ranked by $s(I)$
\end{algorithmic}
\end{algorithm}
Algorithm~\ref{alg:ftf} summarizes the overall inference procedure of the proposed FTF framework.

\subsection{Text-agnostic Recall with Multi-granular Learning}

In the recall stage, the framework encodes each RS image $I$ into a multi-granular visual embedding $\mathbf{v}_1$ without involving any textual query.  
This stage is designed to efficiently narrow down the candidate set from the large gallery while preserving structural and semantic information at multiple scales.  

Formally, each image $I$ is partitioned into coarse and fine patches, denoted as $\mathcal{P}_c$ and $\mathcal{P}_f$, respectively.  
These patches are encoded by separate visual encoders, $E_c$ for coarse and $E_f$ for fine patches, producing token-level embeddings $\mathbf{H}^c$ and $\mathbf{H}^f$:
\begin{equation}
	\mathbf{H}^{g} = E_{g}(\mathrm{PatchTokenize}(I, \mathcal{P}_{g})), \quad g \in \{c,f\}.
	\label{eq:multigranular_tokens}
\end{equation}

The multi-granular tokens are then aggregated to obtain the text-agnostic image embedding $\mathbf{v}_1$:
\begin{equation}
	\mathbf{v}_1 = \mathrm{Agg}(\mathbf{H}^c, \mathbf{H}^f),
	\label{eq:recall_embedding}
\end{equation}
which is subsequently normalized for similarity computation.  
Given a textual query embedding $t_q$, the recall stage efficiently retrieves the top-$K$ candidates:
\begin{equation}
	\mathcal{C} = \mathrm{TopK}\!\left(\left\{ \cos\!\left(t_q, \mathbf{v}_1(I)\right) \right\}_{I \in \mathcal{G}}, K \right),
	\label{eq:topk_candidates}
\end{equation}
where $\mathcal{C}$ is the candidate set passed to the rerank stage.  

This design ensures that $\mathbf{v}_1$ captures both global structures and local details, enabling efficient candidate selection using approximate nearest neighbor search while avoiding expensive text-conditioned interaction over the full gallery.

\subsection{Text-guided Rerank with BTIB}
 
After the recall stage, a small set of candidate images is selected using the text-agnostic embedding $\mathbf{v}_1$. 
Recent studies have shown that structured text--visual interaction can improve fine-grained multimodal matching by guiding textual cues to focus on different semantic patterns in visual features~\cite{24eccvOTSeg}. 
Inspired by this observation, the rerank stage refines the ranking by introducing textual information through the BTIB.

For each candidate image, its coarse and fine tokens, $\mathbf{H}^c$ and $\mathbf{H}^f$, interact with the textual token embeddings $T$ to produce text-guided coarse and fine embeddings, $\mathbf{v}_2$ and $\mathbf{v}_3$:
\begin{equation}
	\mathbf{v}_2, \mathbf{v}_3 = \mathrm{BTIB}(\mathbf{H}^c, \mathbf{H}^f, T).
	\label{eq:btib_outputs}
\end{equation}

The core mechanism of BTIB is a dual normalization process.  
Let the similarity matrix between textual tokens and image tokens be
\begin{equation}
	S = T \mathbf{H}^\top \in \mathbb{R}^{N_t \times N_v},
	\label{eq:similarity_matrix}
\end{equation}
where $T \in \mathbb{R}^{N_t \times d}$ denotes textual token embeddings, and $\mathbf{H} \in \{\mathbf{H}^c, \mathbf{H}^f\}$ denotes coarse or fine image tokens corresponding to $\mathbf{v}_2$ or $\mathbf{v}_3$.  

\textit{Text-side normalization:} apply a softmax along the textual token dimension to ensure fair token allocation:
\begin{equation}
	\tilde{S}_{ij} = \frac{\exp(S_{ij}/\tau)}{\sum_{k=1}^{N_v} \exp(S_{ik}/\tau)}, \quad i=1,\dots,N_t,\; j=1,\dots,N_v,
	\label{eq:text_side_norm}
\end{equation}
where $\tau$ is a temperature hyperparameter.  

\textit{Image-side normalization:} normalize along the image token dimension to prevent token crowding and feature collapse:
\begin{equation}
	\hat{S}_{ij} = \frac{\tilde{S}_{ij}}{\sum_{l=1}^{N_t} \tilde{S}_{lj}}, \quad i=1,\dots,N_t,\; j=1,\dots,N_v.
	\label{eq:image_side_norm}
\end{equation}

The text-guided visual embeddings are obtained via weighted aggregation:
\begin{equation}
	\mathbf{v} = \hat{S}^\top T \in \mathbb{R}^{N_v \times d},
	\label{eq:text_guided_visual_tokens}
\end{equation}
and then pooled along the token dimension to generate coarse and fine embeddings:
\begin{equation}
	\mathbf{v}_2 = \mathrm{Pool}(\mathbf{v}^c), \quad
	\mathbf{v}_3 = \mathrm{Pool}(\mathbf{v}^f),
	\label{eq:pool_v2_v3}
\end{equation}
where pooling outputs a fixed $d$-dimensional embedding for each image. These embeddings $\mathbf{v}_2$ and $\mathbf{v}_3$ are subsequently used in the I$^2$M Loss for joint optimization with $\mathbf{v}_1$. Finally, the rerank stage computes a weighted similarity score for each candidate:
\begin{equation}
	s(I) = (1-\lambda)\cos(t_q, \mathbf{v}_2) + \lambda \cos(t_q, \mathbf{v}_3),
	\label{eq:score_image}
\end{equation}
and ranks the candidate images in $\mathcal{C}$ according to $s(I)$.    
Importantly, BTIB is parameter-free, enabling lightweight semantic refinement.  

\subsection{Inter- and Intra-Modal Loss}

Recent studies have shown that contrastive learning is an effective method for multimodal representation learning and cross-modal retrieval, and the InfoNCE~\cite{21icmlCLIPBASE} objective has been widely used to align matched visual--text pairs while separating non-matching pairs~\cite{22neurocomputingCLIP4Clip,25aclMMRet}. 
Motivated by this line of research, to jointly optimize the recall embedding $\mathbf{v}_1$ and the rerank embeddings $\mathbf{v}_2$ and $\mathbf{v}_3$ under a unified objective, we design an improved Inter- and Intra-Modal (I$^2$M) Loss. Specifically, the proposed loss combines an inter-modal alignment term to enforce cross-modal consistency with an intra-modal structural consistency term to preserve the relative structure of embeddings within each modality:
\begin{equation}
	\mathcal{L}_{\mathrm{I^2M}} = \mathcal{L}_{\mathrm{inter}} + \beta \mathcal{L}_{\mathrm{intra}},
	\label{eq:i2m_loss}
\end{equation}
where $\beta$ balances the two terms.

\textit{Inter-modal alignment.}
The inter-modal term aligns the global text embedding $t_i$ with the three visual embeddings $\mathbf{v}_{1,i}$, $\mathbf{v}_{2,i}$, and $\mathbf{v}_{3,i}$ of the paired image.  
For each branch $k \in \{1,2,3\}$, we compute the pairwise similarity
\begin{equation}
	s_{ij}^{(k)} = \mathbf{t}_i^\top \mathbf{v}_{k,j},
	\label{eq:sim_kij}
\end{equation}
where $\mathbf{v}_{k,j}$ denotes $\mathbf{v}_{1,j}$, $\mathbf{v}_{2,j}$, or $\mathbf{v}_{3,j}$, respectively.  
We then adopt a sigmoid-based alignment objective:
\begin{equation}
	\mathcal{L}_{\mathrm{inter}}^{(k)} =
	\frac{1}{N^2}\sum_{i=1}^{N}\sum_{j=1}^{N}
	\log\left(1+\exp\left(-y_{ij}\, s_{ij}^{(k)}/\tau\right)\right),
	\label{eq:inter_loss_k}
\end{equation}
where $y_{ij}=1$ if $i=j$, and $y_{ij}=-1$ otherwise, and $\tau$ is a temperature hyperparameter.  
The overall inter-modal loss is defined as
\begin{equation}
	\begin{aligned}
		\mathcal{L}_{\mathrm{inter}}
		&= \alpha_1 \mathcal{L}_{\mathrm{inter}}^{(1)}
		+ \alpha_2 \mathcal{L}_{\mathrm{inter}}^{(2)}
		+ \alpha_3 \mathcal{L}_{\mathrm{inter}}^{(3)},
	\end{aligned}
	\label{eq:inter_loss}
\end{equation}
where $\alpha_1$, $\alpha_2$, and $\alpha_3$ control the contributions of the recall and rerank embeddings.  
This term encourages $\mathbf{v}_1$, $\mathbf{v}_2$, and $\mathbf{v}_3$ to maintain consistent semantic alignment with the paired text.

\textit{Intra-modal structural consistency.}
Although $\mathbf{v}_2$ and $\mathbf{v}_3$ are refined by text-guided interaction, they should preserve the intrinsic neighborhood structure established by the text-agnostic embedding $\mathbf{v}_1$.  
To this end, we first construct a batch-level similarity graph based on For $\mathbf{v}_1$, we compute
\begin{equation}
	a_{ij} = \mathbf{v}_{1,i}^\top \mathbf{v}_{1,j}.
	\label{eq:aij_v1}
\end{equation}

For each sample $i$, we select its top-$M$ neighbors $\mathcal{N}_M(i)$ according to $a_{ij}$ and define the normalized neighborhood weights:
\begin{equation}
	g_{ij} =
	\frac{\exp(a_{ij}/\sigma)}
	{\sum_{m \in \mathcal{N}_M(i)} \exp(a_{im}/\sigma)},
	\quad j \in \mathcal{N}_M(i),
	\label{eq:gij}
\end{equation}
where $\sigma$ is a temperature hyperparameter.  
The intra-modal loss then enforces $\mathbf{v}_2$ and $\mathbf{v}_3$ to preserve this local structure:
\begin{equation}
	\begin{aligned}
		\mathcal{L}_{\mathrm{intra}}^{(k)}
		&= \frac{1}{N}\sum_{i=1}^{N}\sum_{j \in \mathcal{N}_M(i)}
		g_{ij}\left(1-\mathbf{v}_{k,i}^\top \mathbf{v}_{k,j}\right),
		\quad k \in \{2,3\}.
	\end{aligned}
	\label{eq:intra_loss_k}
\end{equation}

The final intra-modal term is
\begin{equation}
	\mathcal{L}_{\mathrm{intra}} =
	\mathcal{L}_{\mathrm{intra}}^{(2)} + \mathcal{L}_{\mathrm{intra}}^{(3)}.
	\label{eq:intra_loss}
\end{equation}

By jointly optimizing inter-modal alignment and intra-modal structural consistency, I$^2$M Loss encourages the recall and rerank embeddings to be semantically aligned with the query while maintaining consistent multi-granular representation structure.

\section{Experiments}
\subsection{Experimental Settings}

\textbf{Dataset.} We conduct experiments on the public RS image--text retrieval benchmark RSITMD. This dataset consists of paired RS imagery and textual descriptions, and exhibits high spatial resolution and complex background characteristics.

\textbf{Settings.} We use Transformer-based encoders for both modalities. The text branch adopts a CLIP-style text encoder~\cite{21icmlCLIPBASE} to generate token-level text features and a global sentence-level embedding. The visual branch is built upon a DINOv2-style Vision Transformer~\cite{23arxivDINOv2}, which provides multi-granular image representations for both coarse recall and fine-grained reranking. To ensure a fair evaluation of the proposed framework itself, we do not initialize the model with external pre-trained weights; instead, all parameters are learned from scratch on the target dataset. At inference time, the system follows a two-stage retrieval pipeline: it first performs efficient recall over the full gallery and then applies the proposed fine-grained interaction only to the Top-$K$ candidate set.

\textbf{Evaluation metrics.} We report Recall@K for both image-to-text retrieval (I2T) and text-to-image retrieval (T2I), where $K\in\{1,5,10\}$, and compute mean Recall (mR) as the average of these six values. Let $\mathcal{Q}$ denote the query set, $\mathrm{TopK}(q)$ the top-$K$ retrieved items for query $q$, and $\mathcal{G}(q)$ the set of its ground-truth matched items. Recall@$K$ is defined as
\begin{equation}
	\mathrm{R@}K
	=
	\frac{1}{|\mathcal{Q}|}
	\sum_{q \in \mathcal{Q}}
	\mathbf{1}\!\left(\mathcal{G}(q)\cap \mathrm{TopK}(q)\neq \emptyset\right),
	\label{eq:recall_at_k}
\end{equation}
where $\mathbf{1}(\cdot)$ equals 1 if the condition is true and 0 otherwise. To evaluate retrieval efficiency, we further report latency (Lat.), defined as the average inference time per query, and throughput (Thr.), defined as the number of queries processed per second.

\textbf{Compared methods.} We compare the proposed method with two groups of representative baselines, including task-specific RS image--text retrieval methods and pre-trained VLM-based methods. The former include LW-MCR~\cite{21tgrsLW-MCR}, AMFMN~\cite{22tgrsAMFMN}, KAMCL~\cite{23tgrsKAMCL}, GaLR~\cite{22tgrsGaLR}, PIR~\cite{23acmmmPIR}, FAAMI~\cite{23sensorsFAAMI}, MTGFE~\cite{23RemoteSensingMTGFE}, MSCMAT~\cite{24tgrsMSCMAT}, IERR~\cite{24tgrsIERR}, VGSGN~\cite{24tgrsVGSGN}, SIPN~\cite{25tgrsSIPN}, and FSSN~\cite{25tgrsFSSN}, while the latter include CLIP-Base~\cite{21icmlCLIPBASE}, PETL~\cite{23tgrsPETL}, SkyScript~\cite{24aaaiskysrcipt}, GRAFT~\cite{24iclrGRAFT}, RemoteCLIP~\cite{24tgrsremoteclip}, GeoRSCLIP~\cite{24tgrsGeoRSCLIP}, EBAKER~\cite{24acmmmEBAKER}, and EAPA~\cite{25acmmmEAPA}. For pre-trained methods, we additionally report the scale of pre-training data in Table~\ref{tab:rsitmd_sota_eff} for a more comprehensive comparison.

\subsection{Comparison with State-of-the-Art Methods}

\begin{table*}[t]
	\caption{Comparison of state-of-the-art methods on RSITMD. Methods marked with $\dagger$ use large-scale VLM pre-training. For CLIP-based methods, the reported pre-train data include the original CLIP pre-training data and any additional RS-specific pre-training data when applicable.}
	\label{tab:rsitmd_sota_eff}
	\centering
	\footnotesize
	\renewcommand{\arraystretch}{1.1}
	\resizebox{\textwidth}{!}{%
		\begin{tabular}{lccccccccccc}
			\toprule
			Method & Year & \multicolumn{3}{c}{I2T} & \multicolumn{3}{c}{T2I} & Pre-train data & mR$\uparrow$ & Lat.\,(ms)$\downarrow$ & Thr.\,(q/s)$\uparrow$ \\
			\cmidrule(lr){3-5}
			\cmidrule(lr){6-8}
			& & R@1 & R@5 & R@10 & R@1 & R@5 & R@10 & & & & \\
			\midrule
			
			LW-MCR & 2021 & 4.33 & 13.25 & 19.18 & 3.84 & 14.96 & 25.73 & 0 & 13.55 & - & - \\
			AMFMN  & 2022 & 4.98 & 14.17 & 20.75 & 4.92 & 17.83 & 28.33 & 0 & 15.51 & 150.1 & 6.7 \\
			GaLR   & 2022 & 11.15 & 36.68 & 51.68 & 14.82 & 31.64 & 42.48 & 0 & 31.41 & 291.7 & 3.4 \\
			KAMCL  & 2023 & 6.42 & 18.69 & 36.58 & 7.68 & 22.31 & 36.62 & 0 & 21.38 & - & - \\
			PIR    & 2023 & 18.14 & 41.15 & 52.88 & 12.17 & 41.68 & 63.41 & 0 & 38.24 & 117.6 & 8.5 \\
			FAAMI  & 2023 & 16.15 & 35.62 & 48.89 & 12.96 & 42.39 & 59.96 & 0 & 35.99 & - & - \\
			MTGFE  & 2023 & 16.33 & 48.15 & \underline{66.85} & 17.18 & 40.27 & 53.48 & 0 & 40.38 & - & - \\
			MSCMAT & 2024 & 22.35 & 42.92 & 55.75 & 15.18 & 47.35 & 64.73 & 0 & 41.38 & - & - \\
			IERR   & 2024 & 27.66 & 48.89 & 61.06 & 24.43 & 56.59 & 72.66 & 0 & 48.55 & - & - \\
			VGSGN  & 2024 & 14.16 & 34.96 & 50.66 & 13.23 & 42.57 & 63.43 & 0 & 36.51 & 106.5 & 9.4 \\
			SIPN   & 2025 & 23.01 & 46.02 & 57.96 & 17.04 & 49.87 & 66.95 & 0 & 43.47 & 120.3 & 8.3 \\
			FSSN   & 2025 & \underline{30.53} & 52.21 & 65.49 & 25.71 & 58.32 & \textbf{75.40} & 0 & 51.28 & 128.7 & 7.8 \\
			\midrule
			
			CLIP-Base$^\dagger$  & 2021 & 9.11 & 20.42 & 31.70 & 8.46 & 27.18 & 39.98 & 400M & 22.81 & \textbf{12.0} & \textbf{83.3} \\
			PETL$^\dagger$       & 2023 & 24.16 & 47.12 & 61.28 & 20.41 & 50.53 & 68.54 & 400M & 45.33 & - & - \\
			SkyScript$^\dagger$  & 2024 & 12.05 & 35.74 & 53.68 & 13.81 & 33.14 & 40.48 & 400M+2.6M & 31.48 & 13.5 & 74.1 \\
			GRAFT$^\dagger$      & 2024 & 25.25 & 44.68 & 57.89 & 24.92 & 53.39 & 61.47 & 10.2M & 44.61 & - & - \\
			RemoteCLIP$^\dagger$ & 2024 & 27.88 & 50.66 & 65.71 & 22.17 & 56.46 & 73.41 & 400M+ & 49.38 & \underline{12.5} & \underline{80.0} \\
			GeoRSCLIP$^\dagger$  & 2024 & 30.09 & 51.55 & 63.27 & 23.54 & 57.52 & 74.60 & 400M+5M & 50.10 & 13.0 & 76.9 \\
			EBAKER$^\dagger$     & 2024 & \textbf{34.07} & \textbf{54.20} & \textbf{67.95} & \textbf{28.05} & \textbf{60.35} & \underline{75.31} & 400M+0.2M & \textbf{53.32} & 13.6 & 73.9 \\
			EAPA$^\dagger$       & 2025 & 29.87 & \underline{53.32} & 62.83 & 24.16 & 57.48 & 71.81 & 400M & 49.91 & - & - \\
			\midrule
			
			\textbf{FTF (K=50)}  & 2026 & 27.80 & 49.30 & 63.40 & 24.60 & 55.10 & 71.70 & 0 & 48.65 & 24.7 & 40.6 \\
			\textbf{FTF (K=100)} & 2026 & 29.60 & 51.20 & 65.30 & 26.70 & 57.80 & 72.34 & 0 & 50.49 & 47.7 & 21.0 \\
			\textbf{FTF (K=200)} & 2026 & 30.30 & 52.40 & 66.20 & \underline{27.40} & \underline{59.00} & 73.52 & 0 & \underline{51.47} & 93.7 & 10.7 \\
			\bottomrule
		\end{tabular}%
	}
\end{table*}

Table~\ref{tab:rsitmd_sota_eff} reports the comparison results on RSITMD. Overall, the proposed FTF achieves highly competitive retrieval performance while maintaining a favorable accuracy--efficiency balance. In particular, without using any large-scale pre-training data, FTF consistently outperforms existing task-specific RS image--text retrieval methods in terms of overall retrieval performance. For example, FTF with $K=200$ achieves the best mR of 51.47, surpassing the previous best task-specific method FSSN (51.28). Meanwhile, it achieves the best I2T R@5 of 52.40 and the best T2I R@1 and R@5 of 27.40 and 59.00, respectively, demonstrating the effectiveness of the proposed two-stage retrieval framework.

Compared with pre-trained VLM-based methods, FTF remains competitive even without any external pre-training data. Although methods such as RemoteCLIP and GeoRSCLIP benefit from millions of image--text pairs for pre-training, FTF still achieves better mR than both RemoteCLIP (49.38) and GeoRSCLIP (50.10). These results indicate that the FTF can achieve strong retrieval performance through efficient candidate selection and fine-grained refinement, without relying on large-scale vision--language pre-training.

\subsection{Ablation Study}

To verify the effectiveness of the proposed modules, we conduct ablation experiments on RSITMD under the setting of $K=100$. Specifically, we analyze three key components of FTF, including multi-granular learning, BTIB, and the I$^2$M Loss. For simplicity, we report I2T R@1, T2I R@1, and mR as representative metrics.

\begin{table}[t]
	\caption{Ablation study of FTF on RSITMD with $K=100$.}
	\label{tab:ablation}
	\centering
	\footnotesize
	\renewcommand{\arraystretch}{1.05}
	\begin{tabular}{lccc}
		\toprule
		Variant & I2T R@1 & T2I R@1 & mR \\
		\midrule
		Coarse only & 18.42 & 16.87 & 39.26 \\
		Fine only & 24.83 & 22.94 & 45.31 \\
		w/o BTIB & 27.96 & 25.21 & 46.67 \\
		w/o intra-modal term & 29.08 & 26.11 & 48.72 \\
		Full FTF & 29.60 & 26.70 & 50.49 \\
		\bottomrule
	\end{tabular}
\end{table}

Table~\ref{tab:ablation} reports the ablation results. First, the performance of \emph{Coarse only} and \emph{Fine only} is significantly lower than that of the full model, which confirms the necessity of multi-granular learning. In particular, \emph{Fine only} clearly outperforms \emph{Coarse only}, indicating that fine-grained local details are more important than coarse global structures for RS image--text retrieval. However, the best performance is achieved only when both branches are used together, showing that coarse and fine representations are complementary.

Second, removing BTIB leads to a clear performance drop, with mR decreasing from 50.49 to 46.67. This result demonstrates that the proposed parameter-free text-guided interaction is effective for refining semantic correspondence between RS imagery and textual descriptions. Third, removing the intra-modal term from the I$^2$M Loss also degrades performance, reducing mR from 50.49 to 48.72. This observation shows that, besides inter-modal alignment, preserving the neighborhood structure across multi-granular representations provides additional optimization benefits. 

\subsection{Efficiency–Accuracy balance Analysis}
 
\begin{figure}[!htb]
	\centering
	\includegraphics[width=\columnwidth]{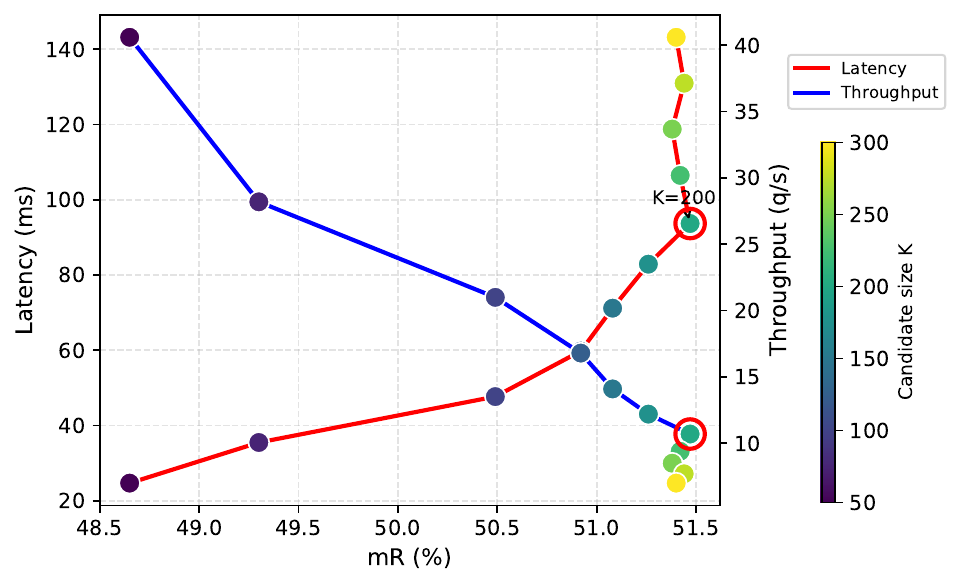}
	\caption{Efficiency--accuracy balance of FTF under different candidate sizes $K$. Single-column plot showing latency and throughput versus mean recall under different candidate sizes K. Point color encodes K, and the figure highlights the balance between retrieval accuracy and efficiency.}
	\label{fig:mR_tradeoff}
\end{figure}

To investigate the impact of the candidate set size $K$ on retrieval performance, we evaluated FTF with $K$ ranging from 50 to 300 on RSITMD.  
Figure~\ref{fig:mR_tradeoff} illustrates the balance between retrieval accuracy (mR) and efficiency metrics, including latency and throughput.  
Each colored dot corresponds to a specific candidate size $K$, and the color map on the right encodes $K$. 

As shown in Figure~\ref{fig:mR_tradeoff}, mR improves steadily as $K$ increases and reaches a plateau around $K=200$.  
Beyond this point, further increasing $K$ provides little gain in accuracy while latency continues to rise and throughput decreases.  
The red and blue dashed lines indicate reference thresholds for latency and throughput, respectively.  
This analysis demonstrates that the proposed two-stage retrieval design effectively balances fine-grained semantic refinement.

\subsection{The Two-stage Retrieval Framework Analysis}
To further verify the advantage of the proposed two-stage design, we compare FTF with a one-stage rerank variant on RSITMD under three candidate sizes, i.e., $K=50$, $K=100$, and $K=200$. The one-stage variant directly performs text-guided reranking, while FTF first conducts text-agnostic recall and then applies reranking to the selected candidates. We report retrieval performance and efficiency in terms of I2T R@1, T2I R@1, mR, latency, and throughput.

\begin{table}[t]
	\caption{Analysis of the two-stage retrieval framework on RSITMD.}
	\label{tab:two_stage}
	\centering
	\footnotesize
	\renewcommand{\arraystretch}{1.08}
	\begin{tabular}{lcccc}
		\toprule
		Variant & mR & Lat. (ms)$\downarrow$ & Thr. (q/s)$\uparrow$ & Speedup$\uparrow$ \\
		\midrule
		Rerank (One stage) & 51.73 & 468.5 & 2.13 & 1.0$\times$ \\
		\midrule
		FTF (Two stage, $K=50$)  & 48.65 & \textbf{24.7} & \textbf{40.6} & \textbf{19.0}$\times$ \\
		FTF (Two stage, $K=100$) & 50.49 & \textbf{47.7} & \textbf{21.0} & \textbf{9.8}$\times$ \\
		FTF (Two stage, $K=200$) & 51.47 & \textbf{93.7} & \textbf{10.7} & \textbf{5.0}$\times$ \\
		\bottomrule
	\end{tabular}
\end{table}

Table~\ref{tab:two_stage} reports the comparison results. Overall, the proposed two-stage framework achieves retrieval performance comparable to one-stage reranking, while substantially improving retrieval efficiency. This is mainly because the expensive text-guided interaction is only applied in the second stage to a small candidate set, instead of the full candidate space. Consequently, FTF reduces the cost of cross-modal interaction and achieves consistent speedup in practice, ranging from about 5$\times$ to nearly 20$\times$ under different candidate sizes. 

Further analysis shows that as the candidate size $K$ increases, the mR of FTF gradually improves and approaches that of the one-stage method, indicating that the first-stage recall is able to preserve most relevant samples.

\subsection{Qualitative Analysis}

\begin{figure}[t!htb]
	\centering
	\includegraphics[width=1.0\linewidth]{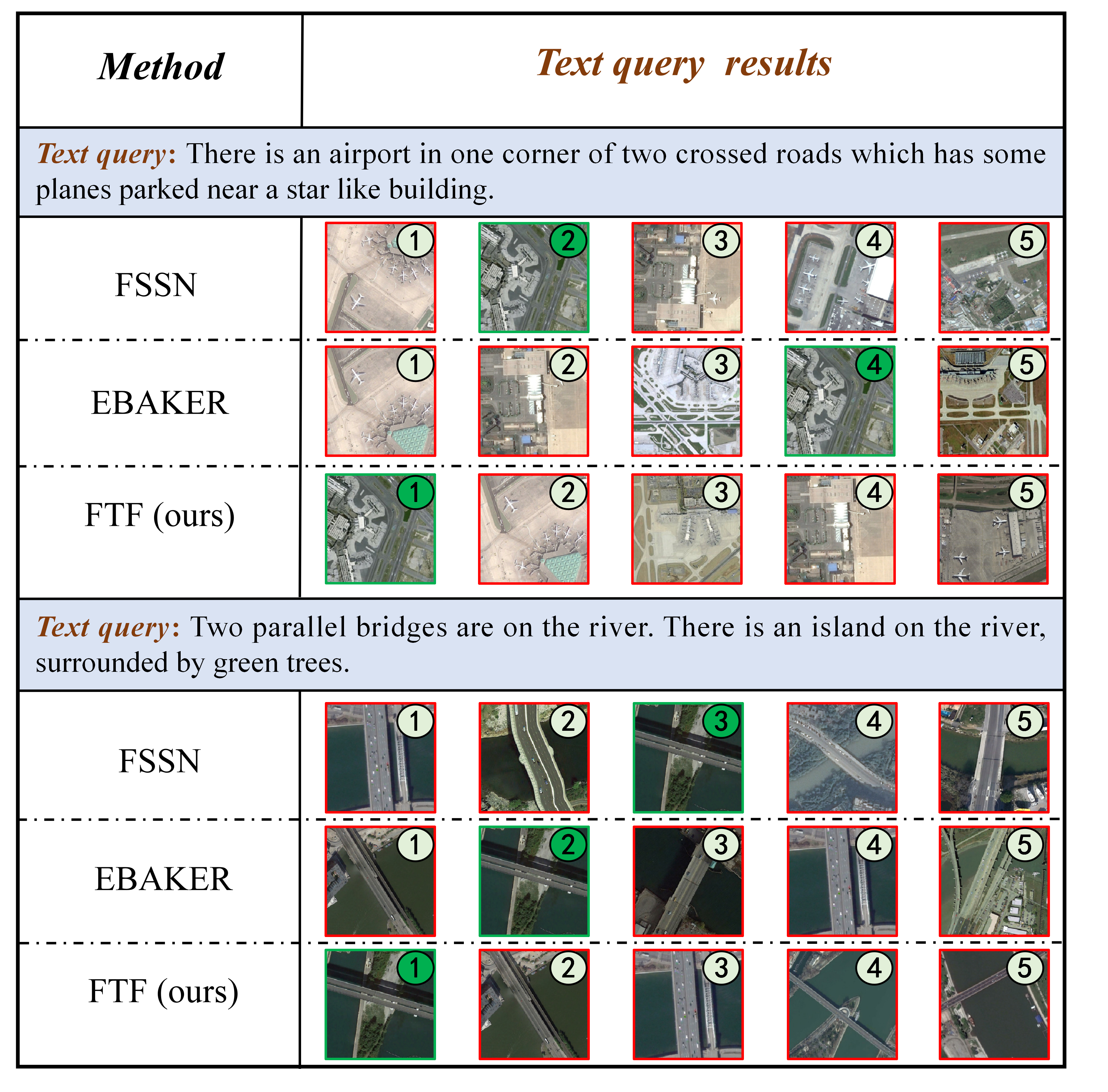}
	\caption{Qualitative comparisons with representative baselines. For each text query, we show the top-5 retrieved results of FSSN, EBAKER, and the proposed method. The numbers denote retrieval ranks, and ground-truth images are highlighted with green boxes. The proposed method retrieves results that are more consistent with both global semantics and fine-grained spatial details.}
	\label{fig:baseline_compare}
\end{figure}

\textbf{Comparison with representative baselines.}
Fig.~\ref{fig:baseline_compare} shows qualitative comparisons with representative baselines under two challenging queries. For the airport query, FSSN and EBAKER retrieve semantically related airport scenes, but fail to rank the ground-truth image at the top because they do not fully capture the detailed spatial constraints. In contrast, the proposed method ranks the correct image at position 1, showing better alignment with both global semantics and fine-grained layout details. For the bridge query, a similar trend can be observed: although the baselines return visually related bridge scenes, their results only partially satisfy the query conditions, whereas our method again ranks the ground-truth image first. These examples show that the proposed method is more effective in handling queries that require both coarse semantic understanding and fine-grained structural discrimination.

\begin{figure*}[!htb]
	\centering
	\includegraphics[width=0.9\linewidth]{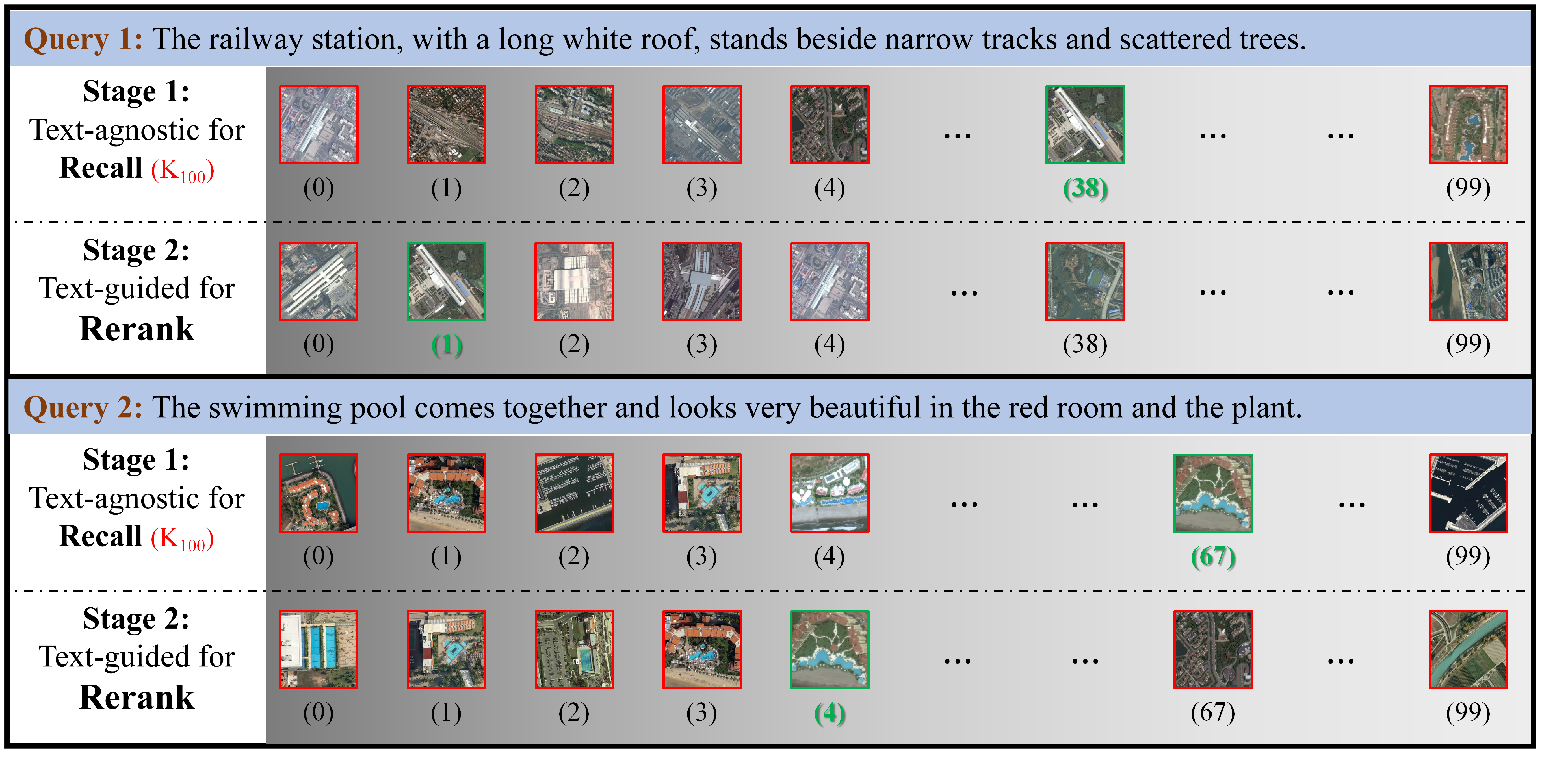}
	\caption{Qualitative examples illustrating the effectiveness of the proposed two-stage retrieval framework. In the recall stage, the ground-truth image is retrieved into the candidate set but may appear at a relatively low rank. After text-guided reranking, the same image is promoted to a much higher position. Ground-truth images are highlighted with green boxes.}
	\label{fig:2stage}
\end{figure*}

\textbf{Effectiveness of the two-stage framework.} Fig.~\ref{fig:2stage} further illustrates the effectiveness of the proposed two-stage retrieval framework. In the first stage, text-agnostic recall is able to retrieve the ground-truth image into the candidate set, but its ranking position may remain relatively low because the retrieval process does not explicitly exploit fine-grained textual cues. In the second stage, text-guided reranking further refines semantic correspondence within the candidate set and significantly improves the ranking of the correct image. For example, in Query 1, the ground-truth image is ranked at position 38 after recall but is promoted to rank 1 after reranking; in Query 2, it moves from rank 67 to rank 4. These examples show that the recall stage effectively preserves semantically relevant candidates, while the rerank stage further improves the final ordering through fine-grained text-guided interaction.

\begin{figure}[!htb]
	\centering
	\resizebox{1.0\columnwidth}{!}{\includegraphics{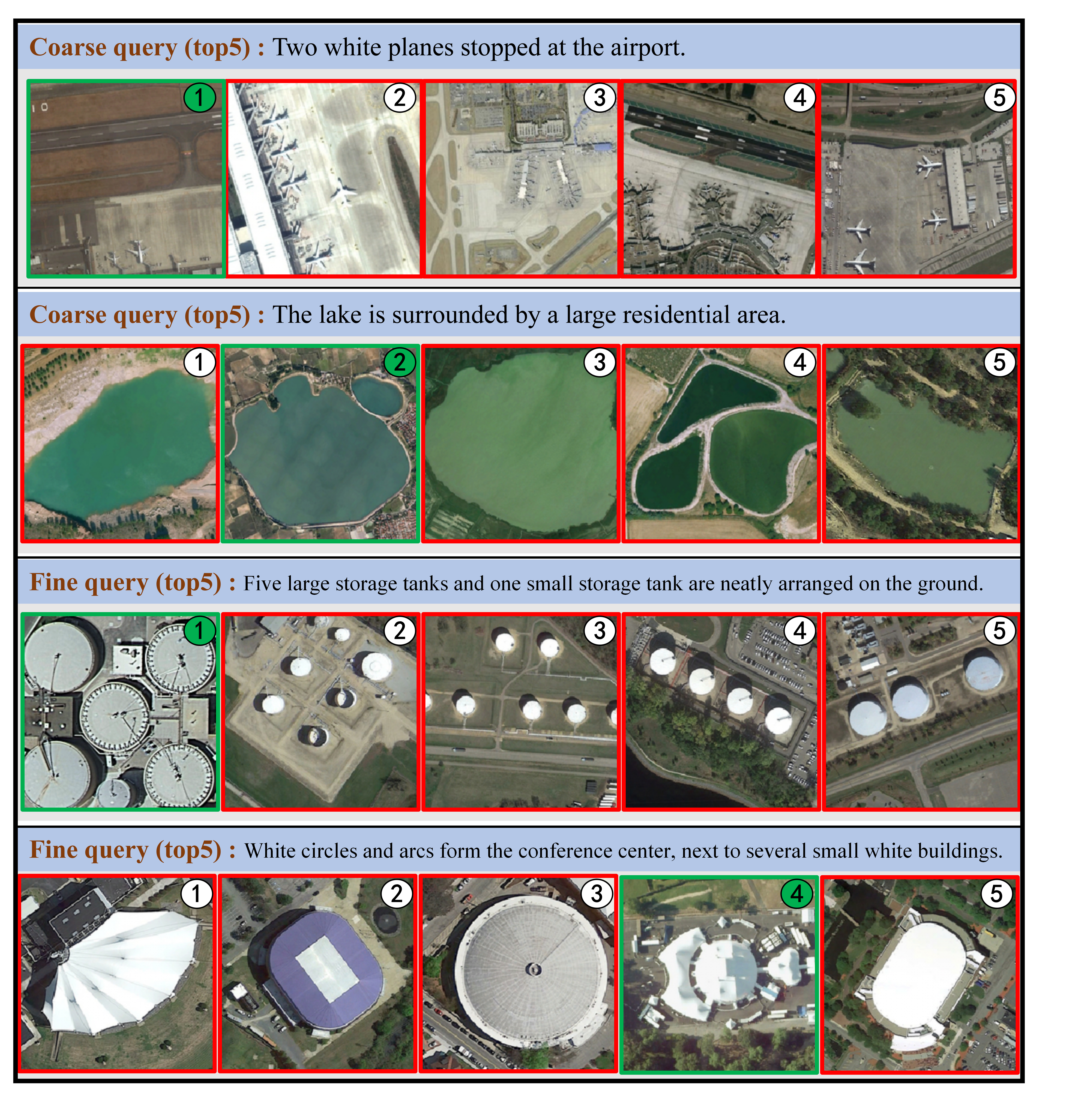}}
	\caption{Qualitative examples illustrating the effectiveness of multi-granular learning. Coarse queries capture global scene structures and dominant semantic categories, while fine queries focus on local objects and structural details. Ground-truth retrievals are highlighted with green boxes.}
	\label{fig:cfquery}
\end{figure}

\textbf{Effectiveness of multi-granular learning.} Fig.~\ref{fig:cfquery} provides qualitative examples demonstrating the benefit of multi-granular learning. Coarse-grained queries capture global scene structures and dominant semantic categories, such as airports or lakes, while fine-grained queries are sensitive to local object arrangements and geometric details, such as the number and layout of storage tanks or the shape of buildings. By combining coarse and fine representations, the framework aligns textual descriptions with RS imagery at multiple semantic levels, effectively handling both large-scale layouts and small-scale objects, consistent with the quantitative improvements reported in Table~\ref{tab:ablation}.

\section{Conclusions}

In this work, we propose FTF, a two-stage framework for RS image--text retrieval that achieves a more favorable balance between retrieval accuracy and efficiency. In the recall stage, a text-agnostic strategy based on multi-granular visual learning is used to efficiently narrow the candidate gallery while preserving both global structures and local details. In the re-rank stage, a text-guided strategy further refines the semantic correspondence between RS imagery and text on the selected candidates. To support this framework, we introduce the parameter-free BTIB module for fine-grained text-guided interaction and the I$^2$M Loss for jointly optimizing inter-modal alignment and intra-modal structural consistency. 

Although the proposed framework shows promising results, it still has several limitations. As a two-stage method, its final re-rank performance depends on the quality of the candidate set produced in the recall stage, and some fine-grained matches may still be missed when the candidate size is very small. 
In addition, while the parameter-free BTIB is efficient, its lightweight design may be less expressive than heavier cross-modal reasoning modules for more complex compositional semantics. In future work, we will explore more robust recall strategies, stronger yet efficient reranking mechanisms, and broader cross-dataset evaluation to further improve the framework.


\bibliographystyle{ACM-Reference-Format}
\bibliography{acmart}

\end{document}